\def\year{2021}\relax
\documentclass[letterpaper]{article} % DO NOT CHANGE THIS
\usepackage{aaai21}  % DO NOT CHANGE THIS
\usepackage{times}  % DO NOT CHANGE THIS
\usepackage{helvet} % DO NOT CHANGE THIS
\usepackage{courier}  % DO NOT CHANGE THIS
\usepackage[hyphens]{url}  % DO NOT CHANGE THIS
\usepackage{graphicx} % DO NOT CHANGE THIS
\usepackage{booktabs}
\usepackage{multirow}
\usepackage{array}
\usepackage{amsmath}
\usepackage{amsfonts}
\usepackage{graphicx}  % DO NOT CHANGE THIS
\usepackage{natbib}  % DO NOT CHANGE THIS AND DO NOT ADD ANY OPTIONS TO IT

\usepackage{caption} % DO NOT CHANGE THIS AND DO NOT ADD ANY OPTIONS TO IT
\usepackage{hyperref}
\makeatletter
\renewcommand\hyper@natlinkbreak[2]{#1}
\makeatother

\newcolumntype{L}[1]{>{\raggedright\let\newline\\\arraybackslash\hspace{0pt}}m{#1}}
\newcolumntype{C}[1]{>{\centering\let\newline\\\arraybackslash\hspace{0pt}}m{#1}}
\newcolumntype{R}[1]{>{\raggedleft\let\newline\\\arraybackslash\hspace{0pt}}m{#1}}
\title{Entity Structure Within and Throughout: Modeling Mention Dependencies\\
 for Document-Level Relation Extraction\\}
\author {
        Benfeng Xu\textsuperscript{\rm 1}\thanks{Work done while the first author was an intern at Baidu Inc..},
        Quan Wang\textsuperscript{\rm 2},
        Yajuan Lyu\textsuperscript{\rm 2},
        Yong Zhu\textsuperscript{\rm 2},
        Zhendong Mao\textsuperscript{\rm 1}\thanks{Corresponding author.} \\
}
\affiliations {
    \textsuperscript{\rm 1} School of Information Science and Technology, University of Science and Technology of China, Hefei, China \\
    \textsuperscript{\rm 2} Baidu Inc., Beijing, China \\
    benfeng@mail.ustc.edu.cn, \{wangquan05, lvyajuan, zhuyong\}@baidu.com, zdmao@ustc.edu.cn \\
}

\begin{document}

\maketitle

\begin{abstract}
Entities, as the essential elements in relation extraction tasks, exhibit certain structure.
In this work, we formulate such structure as distinctive dependencies between mention pairs.
We then propose SSAN, which incorporates these structural dependencies within the standard self-attention mechanism and throughout the overall encoding stage.
Specifically, we design two alternative transformation modules inside each self-attention building block to produce attentive biases so as to adaptively regularize its attention flow.
Our experiments demonstrate the usefulness of the proposed entity structure and the effectiveness of SSAN.
It significantly outperforms competitive baselines, achieving new state-of-the-art results on three popular document-level relation extraction datasets.
We further provide ablation and visualization to show how the entity structure guides the model for better relation extraction. Our code is publicly available.\footnote{\url{https://github.com/PaddlePaddle/Research/tree/master/KG/AAAI2021_SSAN}}
\footnote{\url{https://github.com/BenfengXu/SSAN}}
\end{abstract}

\section{1\quad Introduction}
Relation extraction aims at discovering relational facts from raw texts as structured knowledge.
It is of great importance to many real-world applications such as knowledge base construction, question answering, and biomedical text analysis.
Although early studies mainly limited this problem under an intra-sentence and single entity pair setting, many recent works have made efforts to extend it into document-level texts~\cite{li2016biocreative,yao-etal-2019-docred}, making it a more practical but also more challenging task.

\begin{figure}
\centering
 \includegraphics[width=0.45\textwidth]{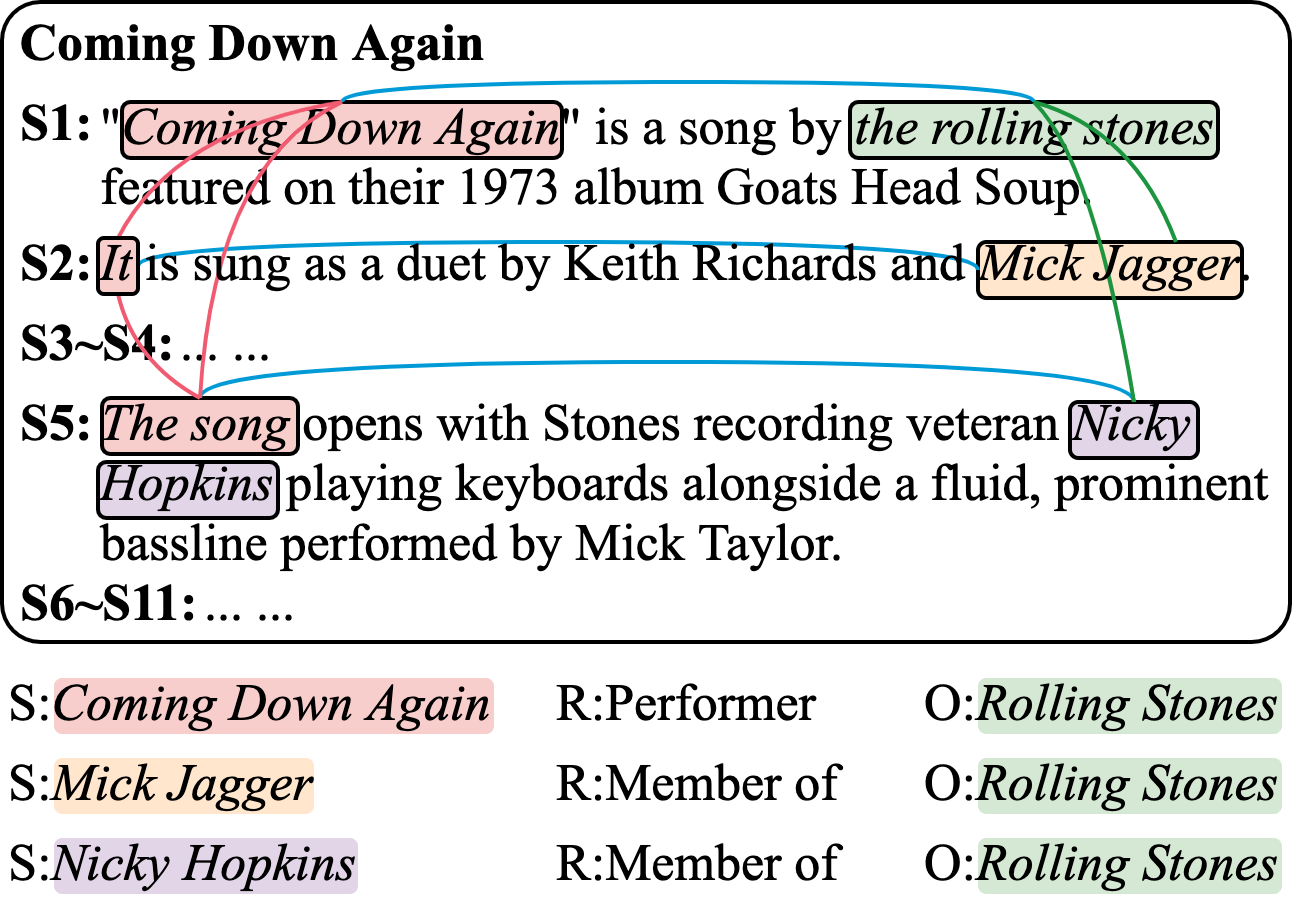}
\caption{
An example excerpted from DocRED. Different mention dependencies are distinguished by colored edges, with the target relations listed in below.}
\label{figure_1}
\end{figure}

Document-level texts entail a large quantity of entities defined over multiple mentions,
which naturally exhibit meaningful dependencies in between.
Figure~\ref{figure_1} gives an example from the recently proposed document-level relation extraction dataset DocRED~\cite{yao-etal-2019-docred}, which illustrates several mention dependencies:
1) \textit{Coming Down Again} and \textit{the Rolling Stones} that both reside in the 1st sentence are closely related, so we can identify \textbf{R1: Performer} (blue link) based on their local context;
2) \textit{Coming Down Again} from the 1st sentence, \textit{It} from the 2nd sentence, and \textit{The song} from the 5th sentence refer to the same entity (red link), so it is necessary to consider and reason with them together;
3) \textit{the Rolling Stones} from the 1st sentence and \textit{Mick Jagger} from the 2nd sentence, though not display direct connections, can be associated via two coreferential mentions: \textit{Coming Down Again} and \textit{it}, which is essential to predict the target relation \textbf{R2: Member of} (green link) between the two entities. Similar dependency also exists between \textit{the Rolling Stones} and \textit{Nicky Hopkins}, which helps identify \textbf{R3: Member of} between them.
Intuitively, such dependencies indicate rich interactions among entity mentions, and thereby provide informative priors for relation extraction.
%Although humans can...
%Use dependency and interaction clearly!

Many previous works have tried to exploit such entity structure, in particular the coreference dependency.
For example, it is a commonly used trick to simply encode coreferential information as extra features, and integrate them into the initial input word embeddings.
~\citet{verga-etal-2018-simultaneously} propose an adapted version of multi-instance learning to aggregate the predictions from coreferential mentions.
Others also directly apply average pooling to the representations of coreferential mentions~\cite{yao-etal-2019-docred}.
In summary, these heuristic techniques only use entity dependencies as complementary evidence in the pre- or post- processing stage, and thus bear limited modeling ability.
Besides, most of them fail to include other meaningful dependencies in addition to coreference.

More recently, graph-based methods have shown great advantage in modeling entity structure~\cite{sahu-etal-2019-inter,christopoulou-etal-2019-connecting,nan-etal-2020-reasoning}. Typically, these methods rely on a general-purpose encoder, usually LSTM, to first obtain contextual representations of an input document. Then they introduce entity structure by constructing a delicately designed graph, where entity representations are updated accordingly through propagation.
%For example,~\cite{luan2019general} designed coreference propagation and relation propagation based on the coreferential structure and coreferential and relational entity structure, which iteratively refine the span representation of entities.
%For example,~\cite{christopoulou-etal-2019-connecting} pool the contextual representation into mention nodes, entity nodes, and sentence nodes, set up a structured graph, then apply an edge-oriented method to iteratively refine the relation representation between target entity pairs.
This kind of approach, however, isolates the context reasoning stage and structure reasoning stage due to the heterogeneity between the encoding network and graph network, which means the contextual representations cannot benefit from structure guidance in the first place.

Instead, we argue that structural dependencies should be incorporated within the encoding network and throughout the overall system.
To this end, we first formulate the aforementioned entity structure under a unified framework, where we define various mention dependencies that cover the interactions in between.
We then propose \textbf{SSAN} (Structured Self-Attention Network), which is equipped with a novel extension of self-attention mechanism~\cite{vaswani2017attention}, to effectively model these dependencies within its building blocks and through all network layers bottom-to-up.
Note that although this paper only focus on entity structure for document-level relation extraction, the method developed here is readily applicable to all kinds of Transformer-based pretrained language models to incorporate any structural dependencies.
%the classic Transformer~\cite{vaswani2017attention} encoder, which is the state-of-the-art architecture and also comes with the benefits of off-the-shelf pretrained language models privilege~\cite{devlin2018bert}.

%Specifically, two alternative modules, namely \textbf{Biaffine Transformation} and \textbf{Decomposed Linear Transformation}, are designed.
%which transform mention dependencies into attentive biases based on their context.
%thus guiding the building up of contextualized and structured entity representations.

To demonstrate the effectiveness of the proposed approach, we conduct comprehensive experiments on DocRED~\cite{yao-etal-2019-docred}, a recently proposed entity-rich document-level relation extraction dataset, as well as two biomedical domain datasets, namely CDR~\cite{li2016biocreative} and GDA~\cite{wu2019renet}.
On all three datasets, we observe consistent and substantial improvements over competitive baselines, and establish the new state-of-the-art.
%along with the 1st position on the DocRED competition leaderboard at the time of submission, surpassing the 2nd by +3.5 Ign F1 score and +3.1 F1 score.
Our contribution can be summarized as follows:
\begin{itemize}
\item We summarize various kinds of mention dependencies exhibited in document-level texts into a unified framework.
By explicitly incorporating such structure within and throughout the encoding network, we are able to perform context reasoning and structure reasoning simultaneously and interactively, which brings substantially improved performance on relation extraction tasks.
\item We propose SSAN that extends the standard self-attention mechanism with structural guidance.
\item We achieve new state-of-the-art results on three document-level relation extraction datasets.
\end{itemize}

\section{2\quad Approach}
This section elaborates on our approach.
We first formalize entity structure in section~\hyperref[2.1:Structure of Entity]{2.1}, then detail the proposed SSAN model in section~\hyperref[2.2:SSAN]{2.2} and section~\hyperref[2.3:Transformation Module]{2.3}, and finally introduce its application to document-level relation extraction in section~\hyperref[2.3:Relation Extraction]{2.4}.

\subsection{2.1\quad Entity Structure}\label{2.1:Structure of Entity}
Entity structure describes the distribution of entity instances over texts and the dependencies among them.
In the specific scenario of document-level texts, we consider the following two structures.\\
\begin{itemize}
\itemsep-1em 
\item{Co-occurrence structure:} Whether or not two mentions reside in the same sentence.\\
\item{Coreference structure:} Whether or not two mentions refer to the same entity.\\
\end{itemize}
Both structures can be described as \textit{True} or \textit{False}.
For \textbf{co-occurrence structure}, we segment documents into sentences, and take them as minimum units that exhibit mention interactions.
So \textit{True} or \textit{False} distinguishes intra-sentential interactions which depend on local context from inter-sentential ones that require cross sentence reasoning.
We denote them as \textit{intra} and \textit{inter} respectively.
For \textbf{coreference structure}, \textit{True} indicates that two mentions refer to the same entity and thus should be investigated and reasoned with together, while \textit{False} implies a pair of distinctive entities that are possibly related under certain predicates.
We denote them as \textit{coref} and \textit{relate} respectively.
%\textit{coreferential} mention pairs indicate the distributed position of each entity, and also bridge other distant entity pairs via multi-hop information propagation, as previously illustrated in figure~\ref{figure_1}, while \textit{non-coreferential} mention pairs entail various relations, and require relational reasoning of models accordingly.
%We argue that these two structures provide informative priors for understanding entities and their relations.
%Intuitively, for coreference structure, mentions of the same entity should share more similarity, and enable multi-hop reasoning by bridging distant entity pairs (as explained in introduction). While mentions of different entities might be related, and the model is expected to perform prediction accordingly.
%We argue that intra-sentence relations and inter-sentence relations require different kinds of reasoning abilities, therefore should be dealt with separately.
In summary, these two structures are mutually orthogonal, resulting in four distinctive and undirected dependencies, as shown in table \ref{table:1}.

\begin{table}[h!]
\centering
\bgroup
\def\arraystretch{1.2}
\begin{tabular}{|ll|c|c|} 
\hline
&&\multicolumn{2}{c|}{Coreference}\\
\cline{3-4}
&& True & False \\
\cline{1-4}
\multicolumn{1}{|c|}{\multirow{2}*{Co-occurence}}&True&\textit{intra+coref}&\textit{intra+relate}\\
\cline{2-4}
\multicolumn{1}{|l|}{}&False&\textit{inter+coref}&\textit{inter+relate}\\
\hline
\end{tabular}
\egroup
\caption{The formulation of entity structure.}
\label{table:1}
\end{table}

\begin{figure*}[t!]
\centering
\includegraphics[width=0.7\textwidth]{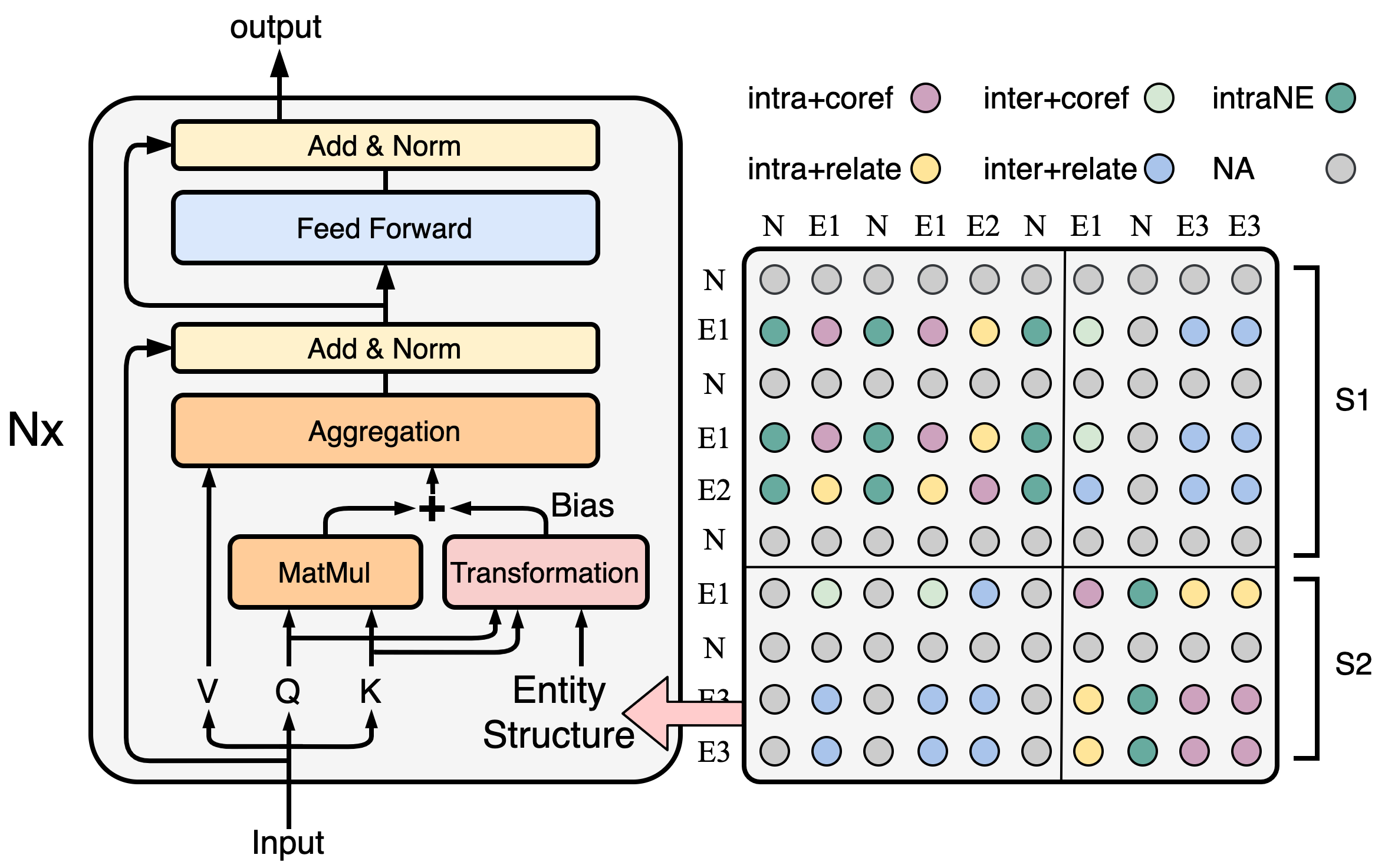}
\caption{The overall architecture of SSAN. Left illustrates structured self-attention as its basic building block. Right explains our entity structure formulation. This minimum example consists of two sentences: $S1$, $S2$, and three entities: $E1$, $E2$ and $E3$. $N$ denotes non-entity tokens. Element in row $i$ and column $j$ represents the dependency from query token $x_i$ to key token $x_j$, we distinguish dependencies using different colors.}
\label{figure_2}
\end{figure*}

Besides the dependencies between entity mentions, we further consider another type of dependency between entity mentions and its intra-sentential \textit{non-entity} (NE) words. We denote it as \textit{intraNE}.
For other inter-sentential non-entity words, we assume there is no crucial dependency, and categorize it as \textit{NA}.
The overall structure is thus formulated into an entity-centric adjacency matrix with all its elements from a finite dependency set: \{\textit{intra+coref}, \textit{inter+coref}, \textit{intra+relate}, \textit{inter+relate}, \textit{intraNE}, \textit{NA}\} (see figure~\ref{figure_2}).

\subsection{2.2\quad SSAN}\label{2.2:SSAN}
SSAN inherits the architecture of Transformer~\cite{vaswani2017attention} encoder, which is a stack of identical building blocks, wrapped up with feedforward network, residual connection, and layer normalization.
As its core component, we propose structured self-attention mechanism with two alternative transformation modules.

Given an input token sequence $x=(x_1, x_2, ..., x_n)$, following the above formulation, we introduce $S=\{s_{ij}\}$ to represent its structure, where $i,j\in\{1,2,...,n\}$ and $s_{ij}\in$\{\textit{intra+coref}, \textit{inter+coref}, \textit{intra+relate}, \textit{inter+relate}, \textit{intraNE}, \textit{NA}\} is a discrete variable denotes the dependency from $x_i$ to $x_j$.
Note that here we extend dependency from mention-level to token-level for practical implementation.
If mention instance consists of multiple subwords ($E3$ in figure~\ref{figure_2}, $S2$), we assign dependencies for each token accordingly.
Within each mention, subword pairs should conform with \textit{intra+coref} and thus are assigned as such.

In each layer $l$, the input representation $\boldsymbol{x}_i^l\in\mathbb{R}^{d_{in}}$ is first projected into query / key / value vector respectively:
\begin{equation}
\boldsymbol{q}_i^l=\boldsymbol{x}_i^l\boldsymbol{W}_l^Q, \boldsymbol{k}_i^l=\boldsymbol{x}_i^l\boldsymbol{W}_l^K, \boldsymbol{v}_i^l=\boldsymbol{x}_i^l\boldsymbol{W}_l^V
\end{equation}
where $\boldsymbol{W}_l^Q$,$\boldsymbol{W}_l^K$,$\boldsymbol{W}_l^V\in\mathbb{R}^{d_{in}\times d_{out}}$.
Based on these inputs and entity structure $S$, we compute unstructured attention score and structured attentive bias, and then aggregate them together to guide the final self-attention flow.

The unstructured attention score is produced by query-key product as in standard self-attention:
\begin{equation}
e_{ij}^l=\frac{\boldsymbol{q}_i^l{\boldsymbol{k}_j^l}^T}{\sqrt{d}}
\end{equation}
Parallel to it, we employ an additional module to model the structural dependency conditioned on their contextualized query / key representations.
We parameterize it as transformations which project $s_{ij}$ along with query vector $\boldsymbol{q}_i^l$ and key vector $\boldsymbol{k}_j^l$ into attentive bias, then impose it upon $e_{ij}^l$:
\begin{equation}
\tilde{e}_{ij}^l=e_{ij}^l+\frac{transformation(\boldsymbol{q}_i^l, \boldsymbol{k}_j^l, s_{ij})}{\sqrt{d}}
\end{equation}
The proposed transformation module regulates the attention flow from $x_i$ to $x_j$.
As a consequence, the model benefits from the guidance of structural dependencies.

After we obtain the regulated attention scores $\tilde{e}_{ij}^l$, a softmax operation is applied, and the value vectors are aggregated accordingly:
\begin{equation}
\boldsymbol{z}_i^{l+1}=\sum_{j=1}^n\frac{exp~\tilde{e}_{ij}^l}{\sum_{k=1}^n exp~\tilde{e}_{ik}^l}\boldsymbol{v}_j^l
\end{equation}
here $\boldsymbol{z}_i^{l+1}\in\mathbb{R}^{d_{out}}$ is the updated contextual representation of $\boldsymbol{x}_i^l$.
Figure~\ref{figure_2} gives the overview of SSAN.
In the next section, we describe the transformation module.
%Other details remain invariant with~\cite{vaswani2017attention}.
%In each layer, different attention heads operate in parallel style, and their results are concatenated as the final output.

\subsection{2.3\quad Transformation Module}\label{2.3:Transformation Module}
To incorporate the discrete structure $s_{ij}$ into an end-to-end trainable deep model, we instantiate each $s_{ij}$ as neural layers with specific parameters, train and apply them in a compositional fashion.
%producing bias scores for every token pair with structural dependency.
As a result, for each input structure $S$ composed of $s_{ij}$, we have a structured model composed of corresponding layer parameters.
As for the specific design of these neural layers, we propose two alternatives: Biaffine Transformation and Decomposed Linear Transformation:
\begin{equation}
\begin{split}
bias_{ij}^l&=Biaffine(s_{ij}, \boldsymbol{q}_i^l, \boldsymbol{k}_j^l)\\
&or\\
&=Decomp(s_{ij}, \boldsymbol{q}_i^l, \boldsymbol{k}_j^l)
\end{split}
\end{equation}
%In order to explicitly model dependencies as discrete variables, we instantiate $s_{ij}$ into corresponding module parameters.

\subsubsection{Biaffine Transformation}
Biaffine Transformation computes the bias as:
\begin{equation}
bias_{ij}^l=\boldsymbol{q}_i^l\boldsymbol{A}_{l, s_{ij}}{\boldsymbol{k}^{l}_j}^T+b_{l, s_{ij}}
\label{equation:6}
\end{equation}
here we parameterize dependency $s_{ij}$ as trainable neural layer $\boldsymbol{A}_{l, s_{ij}}\in\mathbb{R}^{d_{out}\times 1\times d_{out}}$, which attends to the query and key vector simultaneously and directionally, and projects them into a single-dimensional bias.
As for the second term $b_{l, s_{ij}}$, we directly model prior bias for each dependency independent to its context.

\subsubsection{Decomposed Linear Transformation}
Inspired by how~\citet{dai-etal-2019-transformer} decompose the word embedding and position embedding in Transformer, we propose to introduce bias upon query and key vectors respectively, the bias is thus decomposed as:
%\begin{equation}
%\begin{split}
%\tilde{e}_{ij}=&(q_i+\Delta l_q)(k_j+\Delta l_k)^T\\
%=&q_i{k_j}^T+q_i{\Delta l_k}^T+\Delta l_q{k_j}^T+{\Delta l_q}{\Delta l_k}^T
%\end{split}
%\end{equation}
\begin{equation}
bias_{ij}^l=\boldsymbol{q}_i^l\boldsymbol{K}_{l, s_{ij}}^T+\boldsymbol{Q}_{l, s_{ij}}{\boldsymbol{k}_j^l}^T+b_{l, s_{ij}}
\label{equation:7}
\end{equation}
where $\boldsymbol{K}_{l, s_{ij}}$,$\boldsymbol{Q}_{l, s_{ij}}\in\mathbb{R}^d$ are also trainable neural layers.
Intuitively, these three terms respectively represent: 1) bias conditioned on query token representation, 2) bias conditioned on key token representation, and 3) prior bias.

So the overall computation of structured self-attention is:
\begin{equation}
\begin{split}
\tilde{e}_{ij}^l=&\frac{\boldsymbol{q}_i^l{\boldsymbol{k}_j^l}^T+transformation(\boldsymbol{q}_i^l, \boldsymbol{k}_j^l, s_{ij})}{\sqrt{d}}\\
=&\frac{\boldsymbol{q}_i^l{\boldsymbol{k}_j^l}^T+\boldsymbol{q}_i^l\boldsymbol{A}_{l, s_{ij}}{\boldsymbol{k}_j^l}^T+b_{l, s_{ij}}}{\sqrt{d}}\\
or&\\
=&\frac{\boldsymbol{q}_i^l{\boldsymbol{k}_j^l}^T+\boldsymbol{q}_i^l\boldsymbol{K}_{l, s_{ij}}^T+\boldsymbol{Q}_{l, s_{ij}}{\boldsymbol{k}_j^l}^T+b_{l, s_{ij}}}{\sqrt{d}}
\end{split}
\end{equation}
As these transformation layers model structural dependencies adaptively according to context, we do not share them across different layers or different attention heads.

Previously,~\citet{shaw-etal-2018-self} have proposed to model relative position information of input token pair within the Transformer.
They first map the relative distance into embedding, then add them with key vectors before computing the attention score.
Technically, such design can be seen as a simplified version of our Decomposed Linear Transformation, with query conditioned bias only.

\subsection{2.4\quad SSAN for Relation Extraction}\label{2.3:Relation Extraction}
The proposed SSAN model takes document text as input, and builds its contextual representations under the guidance of entity structure within and throughout the overall encoding stage.
In this work, we simply use it for relation extraction with minimum design.
After the encoding stage, we construct a fixed dimensional representation for each target entity via average pooling, which we denote as $\boldsymbol{e}_i\in\mathbb{R}^{d_e}$.
Then, for each entity pair, we compute the probability of relation $r$ from the pre-specified relation schema as:
\begin{equation}
P_r(\boldsymbol{e}_s,\boldsymbol{e}_o)=sigmoid(\boldsymbol{e}_s\boldsymbol{W}_r\boldsymbol{e}_o)
\label{equation_cls}
\end{equation}
where $\boldsymbol{W}_r\in\mathbb{R}^{d_e\times d_e}$.
The model is trained using cross entropy loss:
\begin{equation}
L=\sum_{<s,o>}\sum_{r}CrossEntropy(P_r(\boldsymbol{e}_s,\boldsymbol{e}_o),\overline{y}_r(\boldsymbol{e}_s,\boldsymbol{e}_o))
\label{loss}
\end{equation}
and $\overline{y}$ is the target label.
Given $N$ entities and a relation schema of size $M$, equation~\ref{equation_cls} should be computed $N\times N\times M$ times to give all predictions.
%rely on encoder itself to capture these structures implicitly by itself.
%By utilizing pretrained transformers, we still get an improved understanding of syntax, so do not need help from external tagging tools.
%While the former need intra-sentence structure (e.g., semantic role identifying), the later would require inter-sentence structure (multi-hop reasoning based on the narrative structure).
%These two structures depicts the complexity of this task from two different perspectives and may overlap to form a more complicated structure.

\section{3\quad Experimental Setup}

\subsection{3.1\quad Datasets}
We evaluate the proposed approach on three popular document-level relation extraction datasets, namely DocRED~\cite{yao-etal-2019-docred}, CDR~~\cite{li2016biocreative} and GDA~~\cite{wu2019renet}, all involving challenging relational reasoning over multiple entities across multiple sentences.
We summarize their information in Appendix~\hyperref[appendix:a]{A}.
\subsubsection{DocRED}
DocRED is a large scale dataset constructed from Wikipedia and Wikidata.
It provides comprehensive human annotations including entity mentions, entity types, relational facts, and the corresponding supporting evidence.
There are 97 target relations in total and approximately 26 entities on average in each document.
The data scale is 3053 documents for training, 1000 for development set, and 1000 for test.
Besides, DocRED also collects distantly supervised data for alternative research.
It utilizes a finetuned BERT model to identify entities and link them to Wikidata. Then the relation labels are obtained via distant supervision, producing 101873 document instances at scale.
\subsubsection{CDR}
The Chemical-Disease Reactions dataset is a biomedical dataset constructed using PubMed abstracts.
It contains 1500 human-annotated documents in total that are equally split into training, development, and test sets.
CDR is a binary classification task that aims at identifying induced relation from chemical entity to disease entity, which is of significant importance to biomedical research.
\subsubsection{GDA}
Like CDR, the Gene-Disease Associations dataset is also a binary relation classification task that identify Gene and Disease concepts interactions, but with a much more massive scale constructed by distant supervision using MEDLINE abstracts.
It consists of 29192 documents as the training set and 1000 as the test set.

\subsection{3.2\quad Pretrained Transformers}
%Built upon Transformer, the proposed SSAN model comes with the privilege of leveraging the off-the-shelf pretrained language models, which are mostly developed using the Transformer encoder model.
We initialize SSAN with different pretrained language models including BERT~\cite{devlin-etal-2019-bert}, RoBERTa~\cite{liu2019roberta} and SciBERT~\cite{beltagy-etal-2019-scibert}.
\subsubsection{BERT} BERT is one of the first works that find the success of Transformer in pretraining language models on large scale corpora.
Specifically, it is pretrained using Masked Language Model and Next Sentence Prediction on BooksCorpus and Wikipedia.
BERT is pretrained under two configurations, Base and Large, respectively contains 12 and 24 self-attention layers.
It can be easily finetuned on various downstream tasks, producing competitive baselines.
\subsubsection{RoBERTa} RoBERTa is an optimized version of BERT, which removes the Next Sentence Prediction task and adopts way larger text corpora as well as more training steps.
It is currently one of the superior pretrained language models that outperforms BERT in various downstream NLP tasks.
\subsubsection{SciBERT} SciBERT adopts the same model architecture as BERT, but is trained on scientific text instead.
It demonstrates considerable advantage in a series of scientific domain tasks.
In this paper, we provide SciBERT-initialized SSAN on the two biomedical domain datasets.

\begin{table}[t!]
\begin{tabular*}{0.48\textwidth}{@{}L{3.7cm}C{20mm}C{20mm}@{}}
\toprule
\multirow{2}{*}{\textbf{Model}}&\textbf{Dev}&\textbf{Test}\\
&\textbf{Ign F1} / \textbf{F1}&\textbf{Ign F1} / \textbf{F1}\\
\midrule
ContexAware~(\citeyear{yao-etal-2019-docred})&48.94 / 51.09&48.40 / 50.70\\
EoG$^\ast$(\citeyear{christopoulou-etal-2019-connecting})&45.94 / 52.15&49.48 / 51.82\\
%&&BERT~\citeyear{wang2019fine}&-&54.16&-&53.20\\
BERT Two-Phase~(\citeyear{wang2019fine})&~~~~-~~~~ / 54.42&~~~~-~~~~ / 53.92\\
GloVe+LSR~(\citeyear{nan-etal-2020-reasoning})&48.82 / 55.17&52.15 / 54.18\\
HINBERT~(\citeyear{tang2020hin})&54.29 / 56.31&53.70 / 55.60\\
CorefBERT Base~(\citeyear{ye2020coreferential})&55.32 / 57.51&54.54 / 56.96\\
CorefBERT Large~(\citeyear{ye2020coreferential})&56.73 / 58.88&56.48 / 58.70\\
BERT+LSR~(\citeyear{nan-etal-2020-reasoning})&52.43 / 59.00&56.97 / 59.05\\
CorefRoBERTa~(\citeyear{ye2020coreferential})&57.84 / 59.93&57.68 / 59.91\\
\midrule
\midrule

BERT Base Baseline&56.29 / 58.60&55.08 / 57.54\\
SSAN\textsubscript{Decomp}&56.68 / 58.95&\textbf{56.06} / \textbf{58.41}\\
SSAN\textsubscript{Biaffine}&\textbf{57.03} / \textbf{59.19}&55.84 / 58.16\\
%&&BERT+attentive bias~(Bilinear - \{3\})&\textbf{57.13}&\textbf{59.25}&\textbf{56.51}&\textbf{58.73}\\
\midrule

BERT Large Baseline&58.11 / 60.18&57.91 / 60.03\\
SSAN\textsubscript{Decomp}&58.42 / 60.36&57.97 / 60.01\\
SSAN\textsubscript{Biaffine}&\textbf{59.12} / \textbf{61.09}&\textbf{58.76} / \textbf{60.81}\\
%&&\quad +Distant Data\\
\midrule

RoBERTa Base Baseline&57.47 / 59.52&57.27 / 59.48\\
SSAN\textsubscript{Decomp}&58.29 / 60.22&\textbf{57.72} / 59.75\\
SSAN\textsubscript{Biaffine}&\textbf{58.83} / \textbf{60.89}&57.71 / \textbf{59.94}\\
\midrule
RoBERTa Large Baseline&58.45 / 60.58&58.43 / 60.54\\
SSAN\textsubscript{Decomp}&59.54 / 61.50&59.11 / 61.24\\
SSAN\textsubscript{Biaffine}&\textbf{60.25} / \textbf{62.08}&\textbf{59.47} / \textbf{61.42}\\
\midrule
%\multicolumn{3}{c}{\textit{w/ distant data}}\\
\quad+ Adaptation&\textbf{63.76} / \textbf{65.69}&\textbf{63.78} / \textbf{65.92}\\
\bottomrule
\end{tabular*}
\caption{Results on DocRED. Subscript \textsubscript{Decomp} and \textsubscript{Biaffine} refer to Decomposed Linear Transformation and Biaffine Transformation. Test results are obtained by submitting to official Codalab.
Result with $^\ast$ is from~\citet{nan-etal-2020-reasoning}.}
\label{DocRED}
\end{table}

\begin{table}[t]
\begin{tabular*}{0.48\textwidth}{@{}L{3.3cm}C{9.5mm}C{9.5mm}C{21mm}@{}}
\toprule
\textbf{Model}&\textbf{Dev F1}&\textbf{Test F1}&\textbf{Intra- / Inter- Test F1}\\
\midrule
\cite{gu2017chemical}&-&61.3&57.2 / 11.7\\
BRAN(\citeyear{verga-etal-2018-simultaneously})&-&62.1&- / -\\
CNN+CNNchar(\citeyear{nguyen-verspoor-2018-convolutional})&-&62.3&- / -\\
GCNN(\citeyear{sahu-etal-2019-inter})&57.2&58.6&- / -\\
EoG~(\citeyear{christopoulou-etal-2019-connecting})&63.6&63.6&68.2 / 50.9\\
LSR~(\citeyear{nan-etal-2020-reasoning})&-&61.2&66.2 / 50.3\\
LSR w/o MDP~(\citeyear{nan-etal-2020-reasoning})&-&64.8&68.9 / 53.1\\
BERT~(\citeyear{liu2020document})&-&60.5&- / -\\
SciBERT~(\citeyear{liu2020document})&-&64.0&- / -\\
\midrule
\multicolumn{4}{c}{\textit{methods using external resources}}\\
\fontsize{9.1}{10}\selectfont{\cite{peng2016improving}}&-&63.1&- / -\\
\cite{li2016cidextractor}&-&67.7&58.9 / -~~~~~~~\\
\cite{panyam2018exploiting}&-&60.3&65.1 / 45.7\\
\cite{zheng2018effective}&-&61.5&- / -\\
\midrule
\midrule
BERT Base Baseline&61.7&61.4&69.3 / 44.9\\
SSAN\textsubscript{Decomp}&63.0&61.2&68.6 / \textbf{45.1}\\
SSAN\textsubscript{Biaffine}&\textbf{64.7}&\textbf{62.7}&\textbf{70.4} / 44.7\\
\midrule
BERT Large Baseline&65.3&63.6&70.8 / 49.0\\
SSAN\textsubscript{Decomp}&64.9&64.5&71.2 / 50.2\\
SSAN\textsubscript{Biaffine}&\textbf{65.8}&\textbf{65.3}&\textbf{71.4} / \textbf{52.0}\\
\midrule
SciBERT Baseline&68.2&65.8&71.9 / 53.3\\
SSAN\textsubscript{Decomp}&67.9&67.0&72.6 / 55.8\\
SSAN\textsubscript{Biaffine}&\textbf{68.4}&\textbf{68.7}&\textbf{74.5} / \textbf{56.2}\\
%\midrule
%RoBERTa Large Baseline&63.6&71.7&46.7\\
%SSAN\textsubscript{Biaffine}&64.7&72.5&47.2\\
\bottomrule
\end{tabular*}
\caption{Results on CDR dev set and test set.}
\label{CDR}
\end{table}

\begin{table}[t!]
\centering
\begin{tabular*}{0.48\textwidth}{@{}L{3.2cm}C{10mm}C{10mm}C{21mm}@{}}
\toprule
\textbf{Model}&\textbf{Dev F1}&\textbf{Test F1}&\textbf{Intra- / Inter- Test F1}\\
\midrule
EoG~(\citeyear{christopoulou-etal-2019-connecting})&78.7&81.5&85.2 / 49.3\\
LSR~(\citeyear{nan-etal-2020-reasoning})&-&79.6&83.1 / 49.6\\
LSR w/o MDP~(\citeyear{nan-etal-2020-reasoning})&-&82.2&85.4 / 51.1\\
\midrule
\midrule
BERT Base Baseline&79.8&81.2&84.7 / 60.3\\
SSAN\textsubscript{Decomp}&81.5&\textbf{83.4}&\textbf{86.7} / \textbf{62.3}\\
SSAN\textsubscript{Biaffine}&\textbf{81.6}&82.1&86.1 / 56.8\\
\midrule
BERT Large Baseline&80.4&81.6&84.9 / 61.5\\
SSAN\textsubscript{Decomp}&82.0&83.8&86.6 / \textbf{65.0}\\
SSAN\textsubscript{Biaffine}&\textbf{82.2}&\textbf{83.9}&\textbf{86.9} / 63.9\\
\midrule
SciBERT Baseline&81.4&83.6&\textbf{87.2} / 61.8\\
SSAN\textsubscript{Decomp}&82.5&83.2&87.0 / 60.0\\
SSAN\textsubscript{Biaffine}&\textbf{82.8}&\textbf{83.7}&86.6 / \textbf{65.3}\\
%\midrule
%RoBERTa Large Baseline&81.7&82.7&85.7 / 64.1\\
%SSAN\textsubscript{Biaffine}&82.7&83.9&87.0 / 63.6\\
\bottomrule
\end{tabular*}
\caption{Results on GDA dev set and test set.}
\label{GDA}
\end{table}

\subsection{3.3\quad Implementation Detail}
On each dataset, we give comprehensive results of SSAN initialized with different pretrained language models along with their corresponding baselines for fair comparisons.
The parameters in newly introduced transformation modules are learned from scratch.
All results are obtained using grid search for hyper-parameters (see appendix~\hyperref[appendix:b]{B} for detail) on the development set, then the best model is selected to produce results on the test set.
%(see appendix~\hyperref[appendix:a]{B} for detail)
On DocRED, following the official baseline implementation~\cite{yao-etal-2019-docred}, we utilize naive features including entity type and entity coreference, which is added to the input word embedding. We also concatenate entity relative distance embedding of each entity pair before the final classification.
We preprocess CDR and GDA dataset following~\citet{christopoulou-etal-2019-connecting}. On CDR, after the best hyper-parameter is set, we merge the training set and dev set to train the final model, on GDA, we split 20\% of the training set for development.

\section{4\quad Experiments and Results}
\subsection{4.1\quad DocRED Results}
We conduct comprehensive and comparable experiments on DocRED dataset.
% using four different pretrained transformers (BERT Base/Large and RoBRETa Base/Large) as initialization parameters.
We report both F1 and Ign F1 according to~\citet{yao-etal-2019-docred}. Ign F1 is computed by excluding relational facts that already appeared in the training set.

As shown in table~\ref{DocRED}, SSAN with both \textit{Biaffine} and \textit{Decomp} transformation can consistently outperform their baselines with considerable margin.
In most of the results, \textit{Biaffine} brings more considerable performance gain compared to \textit{Decomp}, which demonstrates that the former is of greater ability to model structural dependencies.
%This also conforms with early related work that applies biaffine classifier in dependency parsing~\cite{DBLP:conf/iclr/DozatM17}.

We compare our model with previous works that either do not consider entity structure or do not explicitly model them within and throughout encoders.
Specifically, ContexAware~\cite{yao-etal-2019-docred}, BERT Two-Phase~\cite{wang2019fine} and HINBERT~\cite{tang2020hin} do not consider the structural dependencies among entities.
EOG~\cite{christopoulou-etal-2019-connecting} and LSR~\cite{nan-etal-2020-reasoning} utilize graph methods to perform structure reasoning, but only after the BiLSTM or BERT encoder.
CorefBERT and CorefRoBERTa~\cite{ye2020coreferential} further pretrain BERT and RoBERTa with a coreference prediction task to enable implicit reasoning of coreference structure.
Results in table~\ref{DocRED} shows that SSAN performs better than these methods. Our best model, SSAN\textsubscript{Biaffine} built upon RoBERTa Large, is \textbf{+2.41} / \textbf{+1.79 Ign F1} better on dev / test set than CorefRoBERTa Large~\cite{ye2020coreferential}, and \textbf{+1.80} / \textbf{+1.04 Ign F1} better than our baseline.
In general, these results demonstrate both the usefulness of entity structure and the effectiveness of SSAN.
%\textbf{+1.8/+1.04 Ign F1} on dev/test than our implemented baseline 

\begin{table}[t!]
\centering
\bgroup
\def\arraystretch{1}
\begin{tabular}{|l|c|c|} 
\hline
\multicolumn{1}{|c|}{\textbf{Dependency}}&\textbf{Ign F1}&\textbf{F1}\\
\hline
SSAN\textsubscript{Biaffine} (RoBERTa Large)&60.25&62.08\\
$-$ \textit{intra+coref}&59.59&61.57\\
$-$ \textit{intra+relate}&59.92&61.91\\
$-$ \textit{inter+coref}&59.87&61.74\\
$-$ \textit{inter+relate}&59.92&61.84\\
$-$ \textit{intraNE}&59.96&61.97\\
$-$ all&58.45&60.58\\
\hline
\end{tabular}
\egroup
\caption{Ablation for entity structure formulation on DocRED dev set. Results when each dependency is excluded, and ``-all’’ degenerates to RoBERTa Large baseline.}
\label{ablation:1}
\end{table}

\begin{table}[!]
\centering
\bgroup
\def\arraystretch{1}
\begin{tabular}{|l|C{1cm}|C{1cm}|} 
\hline
\multicolumn{1}{|c|}{\textbf{Bias Term}}&\textbf{Ign F1}&\textbf{F1}\\
\hline
RoBERTa Large baseline (w/o bias)&58.45&60.58\\
\hline
$+b_{s_{ij}}$&58.62&60.59\\
$+\boldsymbol{Q}_{s_{ij}}\boldsymbol{k}_j^T$&58.79&60.65\\
$+\boldsymbol{q}_i\boldsymbol{K}_{s_{ij}}^T$&59.26&61.31\\
$+\boldsymbol{q}_i\boldsymbol{K}_{s_{ij}}^T+\boldsymbol{Q}_{s_{ij}}\boldsymbol{k}_j^T+b_{s_{ij}}$&59.54&61.50\\
\hline
$+\boldsymbol{q}_i\boldsymbol{A}_{s_{ij}}\boldsymbol{k}_j^T$&59.83&61.75\\
$+\boldsymbol{q}_i\boldsymbol{A}_{s_{ij}}\boldsymbol{k}_j^T+b_{s_{ij}}$&60.25&62.08\\
\hline
\end{tabular}
\egroup
\caption{Ablation for bias terms of two transformation modules on DocRED dev set. Refer to equation~\ref{equation:6} and equation~\ref{equation:7} for specifics, we have removed the layer index $l$ because the ablation is implemented across all layers.}
\label{ablation:2}
\end{table}

Although SSAN is well compatible with pretrained Transformer models, there still exists a distribution gap between parameters in newly introduced transformation layers and those already pretrained ones, thus impedes the improvements of SSAN to a certain extent.
In order to alleviate such distribution deviation, we also utilize the distantly supervised data from DocRED, which shares identical format with the trainset, to first pretrain SSAN before finetuning on the annotated training set for better adaptation. Here we choose our best model, SSAN\textsubscript{Biaffine} built upon RoBERTa Large, and denote it as \textbf{+Adaptation} in table~\ref{DocRED} (see appendix~\hyperref[appendix:b]{B} for hyperparameters setting).
The resulting performance are greatly improved, achieving \textbf{63.78 Ign F1} and \textbf{65.92 F1} on test set as well as the 1st position on the leaderboard\footnote{\url{https://competitions.codalab.org/competitions/20717\#results}} at the time of submission.

\subsection{4.2\quad CDR and GDA Results}
On CDR and GDA datasets, besides BERT, we also adopts SciBERT for its superiority when dealing with biomedical domain texts.
On CDR test set (see Table~\ref{CDR}), SSAN obtains \textbf{+1.3 F1}/\textbf{+1.7 F1} gain based on BERT Base/Large and \textbf{+2.9 F1} gain based on SciBERT, which significantly outperform the baselines and all existing works.
%Moreover, it is also better than methods that utilize external training data~\cite{li2016cidextractor}. 
On GDA (see Table~\ref{GDA}), similar improvements can also be observed.
These results demonstrate the strong applicability and generality of our approach.

\subsection{4.3\quad Ablation Study}
We perform ablation studies of the proposed approach on DocRED. Again, we consider SSAN\textsubscript{Biaffine} built upon RoBERTa Large.
Table~\ref{ablation:1} gives the results of SSAN when each structural dependency is excluded.
It is clear that all five dependencies contribute to the final improvements.
We can arrive at the conclusion that the proposed entity structure formulation is indeed helpful priors for document-level relation extraction.
We can also see that \textit{intra+coref} effects the most among all dependencies.

We also look into the design of two transformation modules by testing each bias term respectively.
As shown in table~\ref{ablation:2}, all bias terms can improve the result over baseline, including the prior bias $+b_{s_{ij}}$ that is only individual values.
Among all bias terms, biaffine bias $+\boldsymbol{q}_i\boldsymbol{A}_{s_{ij}}\boldsymbol{k}_j^T$ is the most effective, brings \textbf{+1.38 Ign F1} improvements solely.
For Decomposed Linear Transformation, key conditioned bias $+\boldsymbol{Q}_{s_{ij}}\boldsymbol{k}_j^T$ produces better results than query conditioned bias $+\boldsymbol{q}_i\boldsymbol{K}_{s_{ij}}^T$, which implies that the key vectors might  be associated with more entity structure information.

\begin{figure*}[t!]
\centering
\includegraphics[width=1\textwidth]{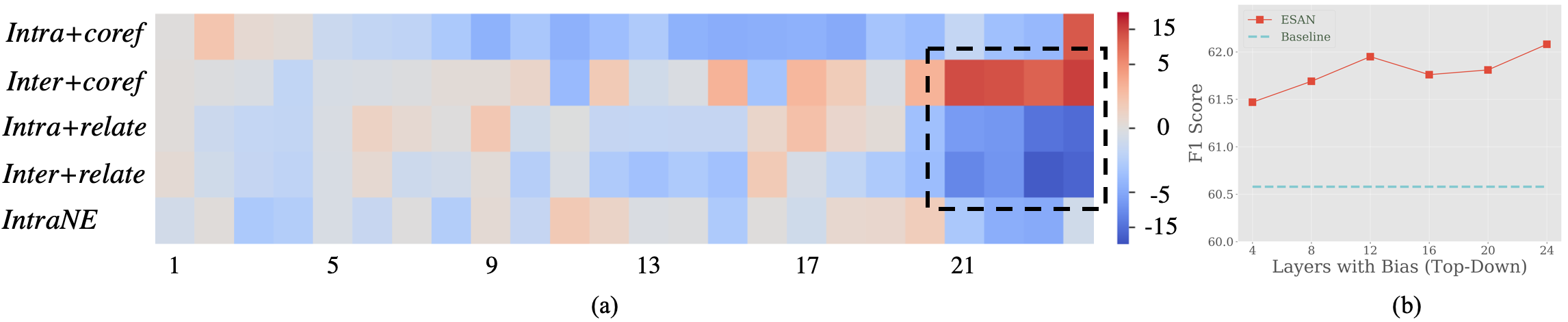}
\caption{(a): Visualization on the learned attentive bias from different layers and different mention dependencies. Results are averaged over the entire dev set and different attention heads. (b): Ablation on number of layers to impose attentive biases.}
\label{figure:3}
\end{figure*}

\subsection{4.4\quad Visualization of Attentive Biases}
As a key feature of SSAN is to formulate entity structure priors into attentive biases, it would be instructive to explore how such attentive biases regulate the propagation of self-attention bottom-to-up.
To this purpose, we collect all attentive biases produced by SSAN\textsubscript{Biaffine} (built upon RoBERTa Large) for DocRED dev instances, categorized according to dependency types, and averaged across all attention heads and all instances.
Figure~\ref{figure:3} (a) is the resultant heatmap, where each cell indicates the value of averaged bias at each layer (horizontal axis) for each entity dependency type (vertical axis).
We can observe meaningful patterns:
1) Along the horizontal axis, the bias is relatively small at bottom layers, where the self-attention score will be mainly decided by unstructured semantic contexts. It then grows gradually and reaches the maximum at the top-most layers, where the self-attention score will be greatly regulated by the structural priors.
2) Along the vertical axis, at the top-most layers (inside the dotted bounding box), bias from \textit{inter+coref} is significantly positive. This conforms with human intuition that coreferential mention pairs might act as a bridge for cross-sentence reasoning, thus should enable more information passing. While biases from \textit{intra+relate} and \textit{inter+relate} appear in contrast.

Based on the discussion, we further investigate the effect of different layers to impose attentive biases. As shown in Figure~\ref{figure:3} (b), with only the top 4 layers (1/6 of the total layers) integrated with entity structure, SSAN can keep +0.89 F1 gain, which confirms that these top-most layers with larger biases indeed impact more significantly. In the meantime, with more layers included, the performance still improves, and reaches the best of +1.50 F1 with all 24 layers equipped with structured self-attention.

\section{5\quad Related Work}
\subsubsection{Document-level RE}
Recent years have seen growing interests for relation extraction beyond single sentence~\cite{quirk-poon-2017-distant,TACL1028}.
Among the most influential works, many have proposed to introduce intra-sentential and inter-sentential syntactic dependencies~\cite{peng2017cross,song-etal-2018-n,gupta2019neural}.
More recently, document-level relation extraction tasks have been proposed~\cite{li2016biocreative,yao-etal-2019-docred}, where the goal is to identify relations of multiple entity pairs from the entire document text, and rich entity interactions are thereby involved.
In order to model these interactions, many graph based methods are proposed~\cite{sahu-etal-2019-inter,christopoulou-etal-2019-connecting,nan-etal-2020-reasoning}.
However, these graph networks are built upon their contextual encoder, which is different from our approach that model entity interactions within and throughout the system.

\subsubsection{Entity Structure}
Entity structure has been shown to be useful in many NLP tasks.
In early works,~\citet{barzilay2008modeling} propose an entity-grid representation for discourse analysis, where the document is summarized into a set of entity transition sequences that record distributional, syntactic, and referential information.
~\citet{ji-etal-2017-dynamic} introduce a set of symbolic variables and state vectors to encode the mentions and their coreference relationships for language modeling task.
~\citet{dhingra-etal-2018-neural} propose Coref-GRU, which incorporates mention coreference information for reading comprehension tasks.
In general, many works have utilized entity structure in various formulation for different tasks.

For document-level relation extraction, entity structure also is essential prior.
For example,~\citet{verga-etal-2018-simultaneously} propose to merge predictions from coreferential mentions.
~\citet{nan-etal-2020-reasoning} propose to model entity interactions via latent structure reasoning.
And ~\citet{christopoulou-etal-2019-connecting} construct a graph of mention nodes, entity nodes, and sentence nodes, then connect them using mention-mention coreference, mention-sentence residency etc., such design provides much more comprehensive entity structure information.
Based on the graph, they further utilize an edge-oriented method to iteratively refine the relation representation between target entity pairs, which is quite different from our approach.
%Actually, there have been several great works investigating how to adapting LSTM architecture with different kinds of structure priors~\cite{DBLP:conf/iclr/KimDHR17,peng2017cross,shen2018ordered}, which mostly entails syntax structure. Instead, in this work, 
%we focus on Transformer~\cite{vaswani2017attention}, the new state-of-the-art architecture that shows powerful capability at modeling long dependencies, and also comes with the benefits of off-the-shelf pretrained language models privilege~\cite{devlin2018bert,liu2019roberta,beltagy2019scibert}.
%Moreover, Transformer can be seen as a variant of densely connected graph network,  and well inherits the latter's potential in modeling structured representation.

\subsubsection{Structured Networks}
Neural networks that incorporate structural priors have been extensively explored.
In previous works, many have investigated how to infuse the tree-like syntax structure into the classical LSTM encoder~\cite{DBLP:conf/iclr/KimDHR17,shen2018ordered,peng2017cross}.
For Transformer encoder, it is also a challenging and thriving research direction.
~\citet{shaw-etal-2018-self} propose to incorporate relative position information of input tokens in the form of attentive bias, which inspired part of this work.
~\citet{wang-etal-2019-extracting} further extend this method to relation extraction task, where the relative position is adjusted into entity-centric form.

%\ifx
\begin{table*}[t!]
\centering
\bgroup
\def\arraystretch{1.2}
\resizebox{0.9\textwidth}{!}{
\begin{tabular}{cc|ccccccc} 
\hline
\multicolumn{2}{c|}{Dataset}&Train&Dev&Test&Entities / Doc&Mentions / Doc&Mention / Sent&Relation\\
\hline
\multirow{2}{*}{DocRED}&Annotated&3053&1000&1000&19.5&26.2&3.58&96\\
&Distant&101873&-&-&19.3&25.1&3.43&96\\
\multicolumn{2}{c|}{CDR}&500&500&500&6.8&19.2&2.48&1\\
\multicolumn{2}{c|}{GDA}&29192&-&1000&4.8&18.5&2.28&1\\
\hline
\end{tabular}}
\egroup
\caption{Summary of DocRED, CDR and GDA datasets. For column \textit{Mention / Sent}, we exclude sentences that do not contain any entity mention. }
\label{datasets}
\end{table*}

\begin{table*}[!]
\centering
\bgroup
\def\arraystretch{1.2}
\resizebox{0.7\textwidth}{!}{
\begin{tabular}{c|ccc|c|cc}
\hline
Dataset&\multicolumn{3}{c|}{DocRED}&CDR&\multicolumn{2}{c}{GDA}\\
\hline
Model&Base&Large&Distant Pretrain&-&Base&Large\\
\hline
learning rate&$5e-5$&$3e-5$&$2e-5$&$5e-5$&$5e-5$&$3e-5$\\
epoch&\multicolumn{2}{c}{$\{40,60,80,100\}$}&$10$&$\{10,20,30,40\}$&\multicolumn{2}{c}{$\{2,4,6\}$}\\
batch size&\multicolumn{3}{c|}{$4$}&$4$&\multicolumn{2}{c}{$\{4,8\}$}\\
\hline
\end{tabular}}
\egroup
\caption{Hyper-parameters Setting.}
\label{hyper-parameters}
\end{table*}
%\fi

\section{6\quad Conclusion and Future Work}
In this work, we formalize entity structure for document-level relation extraction.
Based on it, we propose SSAN to effectively incorporate such structural priors, which performs both contextual reasoning and structure reasoning of entities simultaneously and interactively.
The resulting performance on three datasets demonstrates the usefulness of entity structure and the effectiveness of the SSAN model.

For future works, we give two promising directions: 1) apply SSAN to more tasks such as reading comprehension, where the structure of entities or syntax is useful prior information. 2) extend the entity structure formulation to include more meaningful dependencies, such as more complex interactions based on discourse structure.

\subsection{Acknowledgments}
We thank all anonymous reviewers for their valuable comments. This work is supported by the National Key Research and Development Project of China (No.2018YFB1004300, No.2018AAA0101900), and the National Natural Science Foundation of China (No.61876223, No.U19A2057).

%\ifx
\section{Appendix}
\subsection{A\quad Datasets}\label{appendix:a}
Table~\ref{datasets} details statistics of entities along with other related information of three selected datasets.
We can see that all three datasets entail more than two dozen mentions per document on average, with each sentence contains approximately three mentions on average.
These statistics further demonstrate the complexity of entity structure in document-level relation extraction tasks.

\subsection{B\quad Hyper-parameters Setting}\label{appendix:b}
Table~\ref{hyper-parameters} details our hyper-parameters setting.
All experiment results are obtained using grid search on the development set.
All comparable results share the same search scope.
%\fi

\bigskip

\bibliography{formatting-instructions-latex-2021.bib}
\end{document}